\documentclass[sn-mathphys-num]{sn-jnl}

\usepackage{graphicx}
\usepackage{multirow}
\usepackage{amsmath,amssymb,amsfonts}
\usepackage{amsthm}
\usepackage{mathrsfs}
\usepackage[title]{appendix}
\usepackage{xcolor}
\usepackage{textcomp}
\usepackage{manyfoot}
\usepackage{booktabs}
\usepackage{algorithm}
\usepackage{algorithmicx}
\usepackage{algpseudocode}
\usepackage{listings}

\usepackage[most]{tcolorbox}

\usepackage{color}
\usepackage{soul}
\definecolor{lightyellow}{rgb}{0.97, 0.97, 0}
\sethlcolor{lightyellow}

\usepackage{placeins}

\begin{document}
	
	\title{Semantic Similarity-Informed Bayesian Borrowing for Quantitative Signal Detection of Adverse Events}

	\author[1]{\fnm{François} \sur{Haguinet}}\email{francois.f.haguinet@gsk.com}
	\author[2]{\fnm{Jeffery L.} \sur{Painter}}\email{jeffery.l.painter@gsk.com}
	\author[2]{\fnm{Gregory E.} \sur{Powell}}\email{gregory.e.powell@gsk.com}
	\author[1]{\fnm{Andrea} \sur{Callegaro}}\email{andrea.x.callegaro@gsk.com}
	\author[3,4]{\fnm{Andrew} \sur{Bate}}\email{andrew.x.bate@gsk.com}
	
	\affil*[1]{\orgname{GlaxoSmithKline}, \orgaddress{\city{Wavre}, \country{Belgium}}}
	\affil[2]{\orgname{GlaxoSmithKline}, \orgaddress{\city{Durham}, \state{NC}, \country{USA}}}
	\affil[3]{\orgname{GlaxoSmithKline}, \orgaddress{\city{London}, \country{UK}}}
	\affil[4]{\orgname{London School of Hygiene and Tropical Medicine}, \orgaddress{\city{London}, \country{UK}}}
	
	\abstract{
		
		We present a Bayesian dynamic borrowing (BDB) approach to enhance the quantitative identiﬁcation of adverse events (AEs) in spontaneous reporting systems (SRSs). The method embeds a robust meta-analytic predictive (MAP) prior within a Bayesian hierarchical model and incorporates semantic similarity measures (SSMs) to enable weighted information sharing from clinically similar MedDRA Preferred Terms (PTs) to the target PT. This continuous similarity-based borrowing overcomes limitations of rigid hierarchical grouping in current disproportionality analysis (DPA).\\ 
		
		Using data from the FDA Adverse Event Reporting System (FAERS) between 2015 and 2019, we evaluate our approach - termed IC SSM - against traditional Information Component (IC) analysis and IC with borrowing at the MedDRA high-level group term level (IC HLGT). A reference set (PVLens), derived from FDA product label updates, enabled prospective evaluation of method performance in identifying AEs prior to official labeling.\\
		
		The IC SSM approach demonstrated higher sensitivity (1332/2337=0.570, Youden’s J=0.246) than traditional IC (Se=0.501, J=0.250) and IC HLGT (Se=0.556, J=0.225), consistently identifying more true positives and doing so on average 5 months sooner than traditional IC. Despite a marginally lower aggregate F1-score and Youden’s index, IC SSM showed higher performance in early post-marketing periods or when the detection threshold was raised, providing more stable and relevant alerts than IC HLGT and traditional IC.\\ 
		
		These ﬁndings support the use of SSM-informed Bayesian borrowing as a scalable and context-aware enhancement to traditional DPA methods, with potential for validation across other datasets and exploration of additional similarity metrics and Bayesian strategies using case-level data.	
	}
	
	
	\maketitle	

	\section{Introduction}
	
	Post-marketing safety surveillance of medicines and vaccines during routine healthcare delivery is essential for identifying emerging safety concerns and evaluating risks previously identified or hypothesized during development \cite{RN1}. This process ensures that marketed products continue to reflect an accurate benefit-risk profile. Spontaneous reporting of Individual Case Safety Reports (ICSRs) has long served as the primary data source for detecting new safety signals \cite{RN2}. While clinical review remains fundamental, the increasing volume and complexity of safety data have elevated the role of quantitative methods in pharmacovigilance (PV).
	
	Quantitative analysis of ICSRs was first proposed in the late 1960s and 1970s \cite{RN3, RN4}. Although early applications were sporadic and often limited to specific product issues \cite{RN5, RN6}, the late 1990s marked a turning point, with routine use of these methods driven by increasing data availability and advances in computation \cite{RN7}. While some quantitative signal detection approaches incorporate external data such as sales or prescription volumes to construct background rates \cite{RN8, RN9}, the most widely used methods rely solely on ICSR data due to their practicality and timeliness \cite{RN10,RN42,RN53,RN54,RN55,RN56,RN57}\footnote{It should be noted that sales and prescription data are routinely used in other contexts, such as to establish denominators in PSURs and PBRERs; the distinction here is that they are less commonly used directly in disproportionality-based signal detection.}. The standard methods—introduced in the late 1990s and early 2000s—are based on two-by-two contingency tables and remain the foundation of quantitative signal detection \cite{RN11, RN12, RN13, RN14}, despite their simplicity \cite{RN15}. Although more complex techniques have since been proposed, they have not consistently outperformed these foundational methods in real-world settings \cite{RN16}. Notably, a multi-database comparison from the IMI PROTECT project showed that variation in signal detection performance across methods was often more influenced by the chosen detection threshold than by the specific algorithm employed \cite{RN58}.
	
	Nevertheless, disproportionality analysis (DPA) using these scores has notable limitations. The underlying data are noisy and subject to non-random missingness, and the independent treatment of adverse event (AE) terms limits their ability to capture clinically similar cases that may share an underlying mechanism \cite{RN17}. Addressing this limitation requires a more effective way to cluster related AEs for signal detection.
	
	The Medical Dictionary for Regulatory Activities (MedDRA) \cite{RN18} is an international hierarchical terminology for coding AEs in clinical trials and post-marketing ICSRs. It facilitates consistency in AE classification and enables grouping of related events via its tree structure, which includes levels of specificity from Preferred Terms (PTs) to higher-level groupings. While useful, MedDRA’s rigid hierarchy often fails to capture nuanced clinical relationships between terms \cite{RN19}. For instance, PT Eyelid oedema and PT Periorbital oedema are located far away from each other in the hierarchy. Another example is PT Pneumonitis NOS and PT Pneumonia NOS.
	
	Standard disproportionality methods treat MedDRA PTs independently, unless they are manually grouped under more general terms. This approach can lead to missed signals when related but distinct PTs are analyzed in isolation or to diluted signals when overly broad terms group unrelated AEs. Although general groupings like high-level group terms (HLGTs) or standardized MedDRA queries (SMQs) are sometimes used during post hoc clinical review, there is little evidence that higher-level aggregation improves algorithmic performance \cite{RN20}. As a result, most automated analyses rely on PT or Lower Level Term (LLT) granularity, both of which exist at the same hierarchical level within MedDRA \cite{RN52}.
	
	Bayesian borrowing offers a potential solution to this limitation \cite{RN21}. Previous applications to ICSRs, particularly for vaccines, have demonstrated benefits when borrowing information across PTs within the same System Organ Class (SOC) \cite{RN22}. However, these methods often rely on binary definitions of similarity: PTs are either allowed to borrow from each other or not. This binary framing fails to reflect the continuum of clinical relatedness among AEs. In reality, AEs vary in their informativeness for one another, and a continuous weighting scheme would allow more nuanced borrowing.
	
	Semantic similarity provides a mechanism for introducing such gradation, enabling context-sensitive borrowing across PTs. Although its application in PV remains limited, some related work has explored data-driven clustering of terms that frequently co-occur \cite{RN23, RN24, RN25}. 
	
	Bayesian dynamic borrowing (BDB) has gained traction in clinical trials for integrating historical control data based on similarity to current study data. Pocock \cite{RN26} pioneered this approach, which has since evolved into several frameworks, including the power prior \cite{RN27}, modified power prior \cite{RN28, RN29}, the Meta-Analytic Predictive (MAP) method \cite{RN30}, and the commensurate prior \cite{RN31}. Schmidli et al. \cite{RN32} enhanced the MAP method by introducing a robust MAP prior, incorporating a vague component to mitigate the impact of discordant historical data. 
	
	This manuscript presents a novel adaptation of BDB for spontaneous reporting systems (SRSs), in which semantic similarity is used to guide information sharing across MedDRA PTs. Throughout this work, we refer to this method as IC SSM -- a semantically informed extension of the Information Component (IC) analysis framework. IC SSM embeds a robust MAP prior within a Bayesian hierarchical model, allowing for dynamically weighted borrowing from PTs that are clinically similar to the PT of interest. This continuous, data-driven borrowing approach replaces rigid hierarchical structures with context-sensitive inference and aims to preserve statistical power while reducing noise from clinically irrelevant terms. We evaluate the performance of IC SSM in comparison to both traditional IC analysis and IC analysis using MedDRA HLGT borrowing.
	
	Using data from the FDA Adverse Event Reporting System (FAERS) from 2015 to 2019, we conduct evaluations against a reference set derived from FDA label updates to assess the method’s ability to detect signals prior to official labeling. Sensitivity analyses are also conducted to explore the impact of key model parameters. This work contributes to the advancement of PV methodologies by addressing a long-standing challenge in AE aggregation, aiming to improve the accuracy and timeliness of early signal detection in post-marketing surveillance. 	
	
	\section{Methods}
	
	\subsection{Data}
	
	This study utilized data from FAERS, specifically from the first quarter 2022 release (2022Q1). To mitigate potential confounding effects introduced by the COVID-19 pandemic, all reports submitted after December 31, 2019, were excluded. This time window was selected to align with complementary data sources, such as the Sentinel Initiative \cite{RN33, RN34}, allowing for future comparative analyses using a consistent dataset.
	
	Our analysis focused on a curated cohort of 69 drugs approved by the FDA between January 1, 2015, and December 31, 2016, under submission type 1 (New Molecular Entity, NME). These drugs were identified using our PVLens platform, which processes and integrates Structured Product Labeling (SPL) documents to generate harmonized substance-level identifiers \cite{RN36}. PVLens achieves this by mapping SPL-derived information—such as RxNorm codes and National Drug Codes (NDCs)—into standardized representations, further aligned via the UMLS MTHSPL (FDA Structured Product Labeling Source Information) table to ensure consistent tracking across label versions \cite{RN35}.
	
	To link FAERS reports to these substances, we developed a multi-step mapping strategy. A distinct list of reported product names in FAERS (533,032 entries) was extracted, including both raw reported names and normalized active ingredient fields. This dataset was saved as a CSV file (FAERS\_products.csv) and processed using our FaersMap tool, a component of the PVLens processing pipeline.
	
	The mapping process involved the following steps:
	\begin{enumerate}
		\item Direct string matching of reported or standardized product names to known SNOMED and RxNorm terms.
		\item Substring matching for unresolved terms, excluding ambiguous or overly generic ingredients (e.g., ``aspirin,'' ``vitamin'').
		\item Active moiety matching using normalized ingredient identifiers.
		\item Hierarchical validation to ensure mapped drugs had valid PVLens substance IDs and supporting evidence (e.g., label-based adverse event or indication entries).
	\end{enumerate}
	
	FAERS products that successfully matched to PVLens substances—based on consistent RxNorm or SNOMED identifiers and verified supporting AE or indication data—were retained. From this group, a final set of 69 drugs was selected using two criteria: (1) availability of safety data following FDA approval, and (2) evidence of new safety label content added within two years of product launch. These criteria ensured the inclusion of products with robust post-marketing safety data and documented label changes, enabling evaluation of signal detection performance over time.
	
	All AEs within FAERS were coded using the MedDRA version 26.1. Disproportionality analyses were conducted at the PT level, and semantic similarity measures (SSMs) were computed between PTs using the MedDRA hierarchy.
	
	\subsection{Semantic similarity measures}
	
	Ontology-based SSMs are widely used in biomedical informatics but remain underutilized in PV. These measures enable the quantification of relatedness between medical concepts and support more nuanced grouping of AEs beyond the rigid hierarchical structure of terminologies like MedDRA.
	
	SSMs are generally categorized into four main types:
	\begin{enumerate}
		\item \textbf{Path-based}: Based on shortest path lengths between terms in the ontology graph.
		\item \textbf{Corpus information content based}: Derived from the frequency of terms in a large corpus of biomedical text.
		\item \textbf{Intrinsic IC-based}: Calculated using only the structure of the ontology, without relying on external corpora.
		\item \textbf{Vector-based}: Learned from data, such as co-occurrence patterns in spontaneous reporting systems.
	\end{enumerate}
	
	Among these, intrinsic IC-based SSMs offer particular advantages for safety surveillance, as they do not require external data sources and are robust to updates in terminologies and databases \cite{RN37}.
	
	In our prior evaluation of SSMs for clustering MedDRA PTs in drug safety data, intrinsic IC-based methods consistently outperformed path-based alternatives when measured using F1-scores against SMQs and expert medical review \cite{RN38}.
	
	For this study, we adopted the Sokal measure—an intrinsic IC-based metric derived from the hierarchical integration of MedDRA and SNOMED CT via the Unified Medical Language System (UMLS)—as our preferred semantic similarity method \cite{RN37, RN39}. Several considerations guided the selection of the Sokal measure:
	\begin{itemize}
		\item It achieved among the highest F1-scores (0.403) in our feasibility assessment, comparable to intrinsic Lin \cite{RN40} and intrinsic IC-based Leacock and Chodorow (LCH) \cite{RN41}.
		\item Unlike Lin, the Sokal measure assigns more weight to higher similarity values, which is advantageous in Bayesian borrowing frameworks where accurate weighting enhances inference.
		\item While LCH also performed well, its unbounded scale complicates its use in Bayesian models, which require similarity scores to be normalized between 0 and 1.
	\end{itemize}
	
	Although several intrinsic IC-based SSMs demonstrated strong performance relative to path-based methods, our focus on the Sokal measure was motivated by its favorable balance of discriminative power, interpretability, and compatibility with our BDB framework.
	
	\subsection{Statistical methods}
	
	Table \ref{tab:cont} presents the contingency table used for DPA of each product–event pair, where $y$ denotes the presence of the AE of interest in an individual case safety report (ICSR), and $x$ indicates the presence of the drug of interest in the same ICSR.

	The basic disproportionality measure used in this study is the relative reporting ratio (RRR), also known as the observed-to-expected ratio ($\frac{O}{E}$):
	
	\begin{center}
		$
		\frac{O}{E} = \frac{a}{ \frac{(a + b)(a + c)}{a + b + c + d} }
		$
	\end{center}
	
	This ratio compares the observed count $a$ to the expected count under the assumption of independence between the drug and event.
	
	To evaluate DPA in SRSs, we employed the IC analysis within a Bayesian hierarchical framework. This framework allows the incorporation of a robust MAP prior to enable BDB from other PTs based on their semantic similarity to the PT of interest. The IC SSM approach was compared against two alternatives: (1) standard IC analysis with no borrowing from other PTs, and (2) IC analysis using BDB from PTs within the same MedDRA HLGT, where equal weight was assigned to each PT included in the prior. Across all methods, a statistical signal was defined as a lower limit of the 95\% credibility interval (CI) for the IC exceeding zero. This consistent signal definition allowed us to isolate and assess the added value of semantic similarity-based borrowing in identifying safety signals.
	
	\subsection{Information Component}
	
	The IC is defined as the base-2 logarithm of the ratio of observed to expected joint probabilities of a drug-event pair, under the assumption of independence \cite{RN11}:
	
	\begin{center}
		$
		IC = \log_2 \frac{P(x, y)}{P(x)P(y)}
		$
	\end{center}
	
	Bayesian inference is used to estimate this quantity. Following the approach of Nor{\'e}n et al. \cite{RN42}, we assume that the cell counts $a$, $b$, $c$, and $d$ follow a multinomial distribution:
	\[
	(a, b, c, d) \sim \text{Multinomial}(N; p_a, p_b, p_c, p_d)
	\]
	where $N = a + b + c + d$ is the total number of ICSRs, and $p_a$, $p_b$, $p_c$, and $p_d$ represent the probabilities associated with each cell in the contingency table. A Dirichlet prior is placed on these probabilities:
	\[
	(p_a, p_b, p_c, p_d) \sim \text{Dirichlet}(\alpha_a, \alpha_b, \alpha_c, \alpha_d)
	\]
	resulting in a posterior distribution that is also Dirichlet:
	\[
	\text{Dirichlet}(\gamma_a, \gamma_b, \gamma_c, \gamma_d) \quad \text{with} \quad \gamma_i = \alpha_i + i \quad \text{for} \ i \in \{a, b, c, d\}
	\]
	
	This prior moderates the estimated association strength, particularly when data support is weak, helping to stabilize estimates and address zero-cell issues. It is configured to center the posterior IC distribution between zero and the observed $log(O/E)$, thereby providing a conservative yet robust inference approach.
	
	In this study, posterior estimates of the IC and its credible interval were derived from Monte Carlo simulations. As the IC distribution is typically unimodal, the posterior mean estimate (PME) was used as the central value. Although Nor{\'e}n recommends using the maximum a posteriori  estimate, we opted for the PME due to its conservative nature. This choice has limited influence on signal detection when alerts are defined by the CI lower limit but is important in the context of Bayesian borrowing, where both the central IC and its variance are used to inform the prior.
	
	All IC calculations were performed using the publicly available \texttt{BCPNN} R function\footnote{\url{https://github.com/bips-hb/pvm/blob/master/R/BCPNN.R}} \cite{RN15}.

	\subsection{Robust MAP prior for BDB based on semantic similarity between outcomes}
	
	The IC SSM method is inspired by dynamic borrowing techniques used in clinical trials, particularly the robust MAP prior framework \cite{RN30,RN32}. In our context, we adapt this approach for PV by dynamically borrowing information from other AEs reported with the same product. specifically, for each product–event pair of interest, we construct a robustified MAP prior using a weighted combination of IC estimates from similar events, leveraging SSMs between MedDRA PTs.
	
	Let $Y_{\mathrm{I}}$ and $IC_{\mathrm{I}}$ denote the data and IC for the product–event pair of interest. Let $Y_{\mathrm{B}} = (Y_1, \ldots, Y_S)$ and $IC_{\mathrm{B}} = (IC_1, \ldots, IC_S)$ represent the data and IC values of $S$ semantically similar events with SSMs above a prespecified threshold, and let $SSM_{\mathrm{B}} = (SSM_1, \ldots, SSM_S)$ denote their corresponding similarity scores, where each $SSM_s \in [0, 1]$, with 0 indicating no similarity and 1 indicating synonymy. Inference on $IC_{\mathrm{I}}$ is based on both direct evidence from $Y_{\mathrm{I}}$ and indirect evidence from $(Y_{\mathrm{B}}, SSM_{\mathrm{B}})$, implemented in two steps. first, a MAP prior $p(IC_{\mathrm{I}} \mid Y_{\mathrm{B}}, SSM_{\mathrm{B}})$ is constructed. Then, this prior is updated via Bayes' theorem:
	
	\[
	p(IC_{\mathrm{I}} \mid Y_{\mathrm{I}}, Y_{\mathrm{B}}, SSM_{\mathrm{B}}) \propto p(Y_{\mathrm{I}} \mid IC_{\mathrm{I}}) \cdot p(IC_{\mathrm{I}} \mid Y_{\mathrm{B}}, SSM_{\mathrm{B}})
	\]
	
	Assuming each estimated $\widehat{\mathrm{IC}}_s$ follows a normal distribution centered on its true value:
	
	\[
	\widehat{\mathrm{IC}}_s \mid \mathrm{IC}_s, \mathrm{VIC}_s \sim \mathcal{N}(\mathrm{IC}_s, \mathrm{VIC}_s), \quad s = 1, \ldots, S
	\]
	
	we account for heterogeneity among similar events using a random-effects framework:
	
	\[
	IC_{\mathrm{I}}, IC_1, \ldots, IC_S \sim \mathcal{N}(\mu, \tau^2)
	\]
	
	Here, $\mu$ is the population mean and $\tau$ the between-event standard deviation.
	
	In standard MAP estimation, IC values are typically weighted by the inverse of their variances. In our adaptation, these weights are further modified by semantic similarity to reflect both statistical uncertainty (via variance) and clinical relevance (via semantic similarity). This approach is analogous to meta-analytic techniques that incorporate study quality into weighting schemes \cite{RN43}.
	
	To illustrate the role of SSMs in the weighting process, we first present the fixed-effect meta-analytic formulation of the MAP prior:
	
	\[
	\widehat{p}_{\mathrm{B}}(\mathrm{IC}_{\mathrm{I}}) \sim \mathcal{N}(\widehat{\mu}_m, \widehat{V}_m)
	\]
	
	with the weighted mean and variance defined as:
	
	\[
	\widehat{\mu}_m = \frac{ \sum_s \left( \frac{\mathrm{SSM}_s}{\mathrm{VIC}_s} \cdot \mathrm{IC}_s \right) }{ \sum_s \frac{ \mathrm{SSM}_s }{ \mathrm{VIC}_s } }
	\quad \text{and} \quad
	\widehat{V}_m = \frac{ \sum_s \frac{ \mathrm{SSM}_s^2 }{ \mathrm{VIC}_s } }{ \left( \sum_s \frac{ \mathrm{SSM}_s }{ \mathrm{VIC}_s } \right)^2 }
	\]
	
	Although this fixed-effect formulation clarifies how SSMs are incorporated, our implementation used a random-effects meta-analysis to accommodate heterogeneity. MAP priors were estimated using restricted maximum likelihood via the \texttt{rma()} function in the R package \texttt{metafor} \cite{RN44}.
	
	Even when SSMs are derived from structured and clinically meaningful ontologies, the assumption of exchangeability among the borrowed PTs may not fully hold. Such violations can introduce prior–data conflict. To mitigate this, we adopt a robust version of the MAP prior:
	
	\[
	\widehat{p}_{\mathrm{BR}}(\mathrm{IC}_{\mathrm{I}}) = w \cdot \widehat{p}_{\mathrm{B}}(\mathrm{IC}_{\mathrm{I}}) + (1 - w) \cdot p_{\mathrm{V}}(\mathrm{IC}_{\mathrm{I}})
	\]
	
	where $\widehat{p}_{\mathrm{B}}(\mathrm{IC}_{\mathrm{I}})$ is the MAP prior from the meta-analysis, and $p_{\mathrm{V}}(\mathrm{IC}_{\mathrm{I}})$ is a vague normal prior centered at 0, representing the null hypothesis of no association. This formulation provides robustness when the IC of the event of interest diverges meaningfully from those of similar events, even after weighting. It aligns with Bayesian DPA strategies that emphasize shrinkage toward zero in the presence of weak evidence \cite{RN7,RN10,RN11,RN42}.
	
	Following Schmidli et al. \cite{RN32}, the vague prior was based on the log-transformed ratio of a binomial proportion with a Beta$(1,1)$ prior. The resulting vague prior for the IC is expressed as:
	
	\[
	\log_2 \left( \frac{\mathrm{Beta}(1,1)}{\mathrm{Beta}(1,1)} \right)
	\]
	
	which corresponds to a standard deviation of approximately 2.0, capturing substantial uncertainty.
	
	While $w$ could be fixed arbitrarily (e.g., $w = 0.8$), such an approach fails to account for the level of similarity between the PT of interest and the PTs used for borrowing. Importantly, even if relative similarities are incorporated in the MAP prior, a fixed $w$ does not distinguish between borrowing from PTs with uniformly low similarity and borrowing where at least one PT has a similarity score near 0.95.
	
	To reflect this nuance, we link $w$ to the maximum observed similarity $\max(SSM_{\mathrm{B}})$, so that borrowing is more substantial when at least one PT is highly similar. If a single PT has strong similarity to the PT of interest, it should dominate the MAP prior—even if other borrowed terms are less related. In contrast, if all SSMs are low, borrowing should be limited. Using the maximum rather than the mean or median ensures that $w$ reflects the strongest available similarity.
	
	Prior work has shown that including only PTs with SSM $> 0.3$ improves clustering accuracy and alignment with clinical expert judgment \cite{RN38}. This threshold also enhances computational efficiency in large-scale DPA.
	
	The resulting robust prior is a mixture of two conjugate normal distributions:
	
	\[
	\widehat{p}_{\mathrm{BR}}(\mathrm{IC}_{\mathrm{I}}) 
	= w \cdot \mathcal{N}(\mathrm{IC}_{\mathrm{I}} \mid \widehat{\mathrm{IC}}_{\mathrm{B}}, \widehat{\mathrm{VIC}}_{\mathrm{B}})
	+ (1 - w) \cdot \mathcal{N}(\mathrm{IC}_{\mathrm{I}} \mid 0, 2^2)
	\]
	
	and the posterior distribution, also a mixture, is given by:
	
	\begin{align*}
		\widehat{p}_{\mathrm{BR}}(\mathrm{IC}_{\mathrm{I}} \mid Y_{\mathrm{I}}) =\; & \widetilde{w} \cdot \mathcal{N} \left( 
		\mathrm{IC}_{\mathrm{I}} \,\middle|\, 
		\frac{ 
			\widehat{\mathrm{IC}}_{\mathrm{B}} \, \widehat{\mathrm{VIC}}_{\mathrm{B}}^{-1} 
			+ \widehat{\mathrm{IC}}_{\mathrm{I}} \, \widehat{\mathrm{VIC}}_{\mathrm{I}}^{-1} 
		}{ 
			\widehat{\mathrm{VIC}}_{\mathrm{B}}^{-1} + \widehat{\mathrm{VIC}}_{\mathrm{I}}^{-1} 
		}, 
		\frac{1}{ 
			\widehat{\mathrm{VIC}}_{\mathrm{B}}^{-1} + \widehat{\mathrm{VIC}}_{\mathrm{I}}^{-1} 
		} 
		\right) \\
		& + (1 - \widetilde{w}) \cdot \mathcal{N} \left( 
		\mathrm{IC}_{\mathrm{I}} \,\middle|\, 
		\frac{ 
			\widehat{\mathrm{IC}}_{\mathrm{I}} \, \widehat{\mathrm{VIC}}_{\mathrm{I}}^{-1} 
		}{ 
			4^{-1} + \widehat{\mathrm{VIC}}_{\mathrm{I}}^{-1} 
		}, 
		\frac{1}{ 
			4^{-1} + \widehat{\mathrm{VIC}}_{\mathrm{I}}^{-1} 
		} 
		\right)
	\end{align*}
	
	where $\widetilde{w}$ is the updated posterior mixture weight, computed according to Best et al. \cite{RN45}.

	\subsection{Reference Set}
	
	Robust evaluation of signal detection methods in PV requires high quality references sets \cite{RN36}. Nor{\'e}n et al. (2014) emphasized the importance of assessing methods using emerging rather than established adverse drug reactions (ADRs), underscoring the need for time-stamped reference standards that reflect the state of knowledge before a given label change \cite{RN46}. While several initiatives have attempted to construct large-scale reference sets over the past decade, including those based on electronic health records or spontaneous reports, consistent quality and temporal alignment have remained elusive \cite{RN47}.
	
	To address this gap, we utilized a newly developed reference set generated using PVLens, an automated system that extracts indications, black box warnings, and AE terms from FDA Structured Product Labeling (SPL) documents and maps them to MedDRA \cite{RN36}. PVLens integrates RxNorm, SNOMED CT, and the UMLS MTHSPL tables to align AE terms with their respective substances and track label changes over time. Its automated NLP pipeline ensures high recall of labeled safety events, and expert adjudication has demonstrated reliable precision in its ability to map and assign MedDRA terms.
	
	For this study, we constructed a positive control set consisting of MedDRA PTs added to U.S. product labels between 2016 and 2019, capturing newly labeled safety information after initial FDA approval. These terms were extracted directly from SPL change logs and verified against historical label versions. By anchoring each AE to its date of labeling, we ensured temporal separation between the data used for signal detection (2015–2019 FAERS) and the subsequent inclusion of that AE in the label, enabling a prospective evaluation of method performance and mitigating reporting bias.
	
	For the negative control set, we followed the guidance of Nor{\'e}n et al. by including all PTs that did not share the same MedDRA High-Level Term (HLT) as any of the positive control PTs, regardless of whether they appeared in historical labeling \cite{RN46}. This conservative definition helps reduce the risk of inadvertently labeling true signals as negatives while maintaining a realistic class imbalance that reflects the distribution of potential safety signals in real-world data.
	
	By leveraging PVLens as a validated, time-aware reference source, this approach improves upon past methods by enabling reproducible, expert-informed evaluations of safety signal detection across dynamic PV datasets.
	
	\subsection{Performance evaluations}			
	
	Performance evaluations were conducted on a quarterly basis from January 1, 2016, to December 31, 2019. For each quarter, we computed sensitivity, specificity, positive predictive value (PPV), Youden's index, and F1-score using the data accumulated since January 1, 2015, up to that point. These metrics were chosen to offer a multifaceted assessment of the methods' effectiveness in signal detection, balancing the trade-offs between the sensitivities, the false positive rates and the precisions. The detailed quarterly analysis allowed for the identification of trends and potential difference in the relative performance of the PV methods under investigation when data accumulates, ensuring a robust evaluation framework that aligns with the dynamic nature of emerging safety signals.
	
	In addition, performance metrics were summarized across the entire analysis period to provide a holistic view of the methods' effectiveness. definitions were as follows: true positives (TP) were defined as positive controls alerted before the label update, with positive controls alerted after the label update being ignored; false positives (FP) were negative controls alerted at any time; true negatives (TN) were negative controls never alerted; and false negatives (FN) were positive controls never alerted before the label update.
	
	Descriptive analyses were conducted to compare the performance of two methods using a one-to-one comparison framework. specifically, we categorized the detection outcomes to detail the number of TPs uniquely identified by each method, the number of TPs concurrently detected by both methods, and the number of TPs that were not detected by either method. Furthermore, we examined the differences in the time-to-detection of the TPs. For TPs detected by only one of the two methods, 2020Q1 was imputed as the quarter of detection.
	
	The performance of the IC SSM is contingent upon three key parameters: the prior value of the weight of the MAP prior (w) in the robust mixed prior, the standard deviation of the vague prior ($\sigma$), and the lower limit of the SSM (minSSM) for PTs to be incorporated into the MAP prior. The uncertainty about the best combination of these parameters led us to perform some sensitivity analysis. To evaluate the influence of these parameters, we conducted a series of analyses varying one parameter at a time, keeping the other two as in a reference parametrization with $\sigma = 2.0$, $w=max(SSM)$ and $minSSM=0.3$.

	\section{Results}	
	
	\subsection{Temporal Trends in Method Performance}
	
\begin{figure}
		\centering
		\fbox{\includegraphics[width=0.95\textwidth]{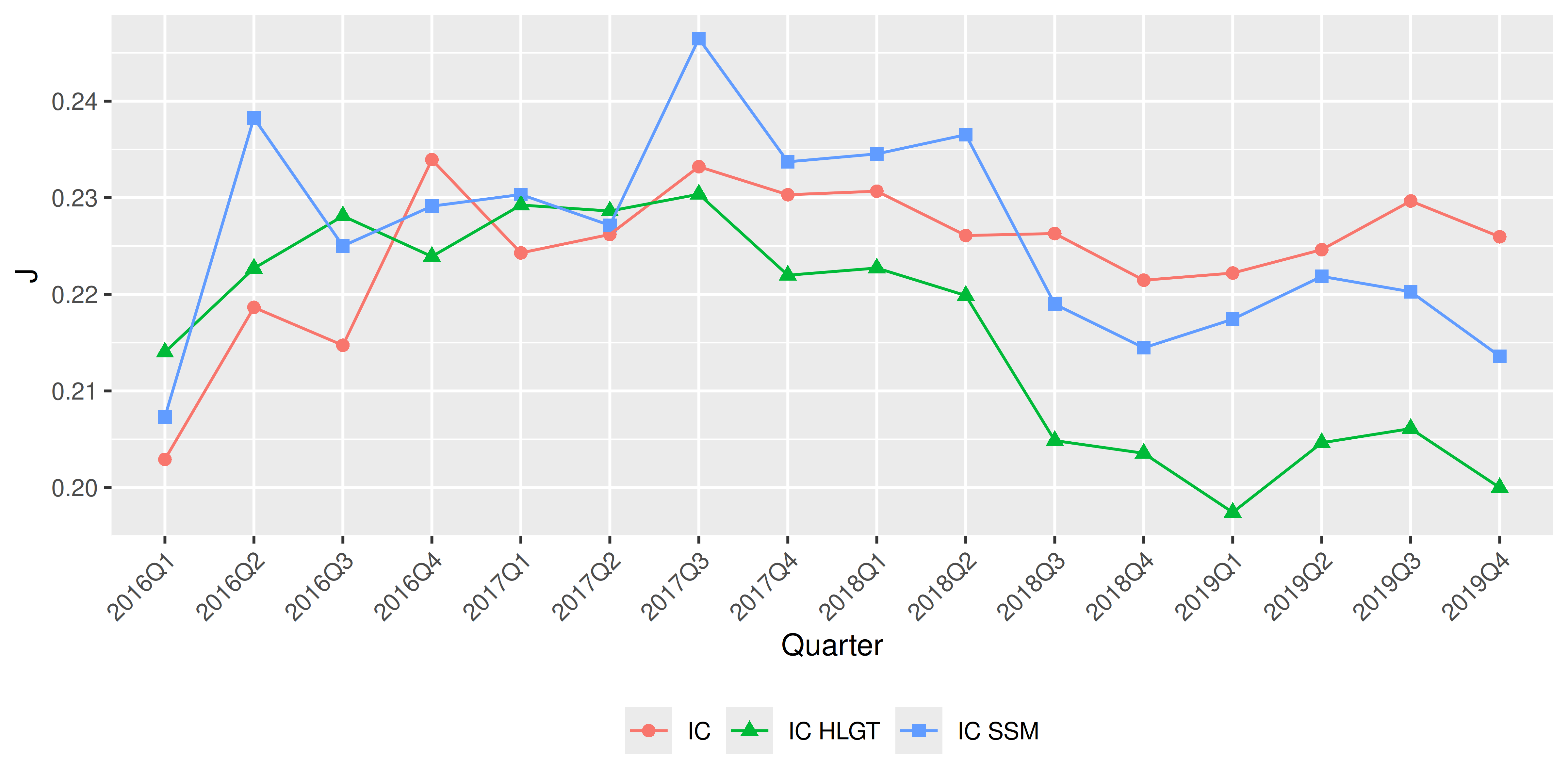}}
		\caption{Evolution over time of the Youden's index (J)}
		\label{fig:fig_01}
	\end{figure}

	The performance of the proposed IC SSM method was evaluated over time to assess its practical value in dynamic post-marketing surveillance scenarios. IC SSM demonstrated clear advantages in sensitivity and early detection -- two key priorities in PV, where timely hypothesis generation can mitigate further patient exposure. The associated trade-off with specificity was favorable during the early post-marketing period, when many safety issues remained unknown. This advantage diminished later, as the pool of unidentified positive controls decreased. These dynamics are reflected in the evolution of Youden’s index over time (Figure \ref{fig:fig_01}).
	
	During the study period, the number of distinct product-event pairs reported per quarter increased markedly, from 3,474 in 2016Q1 to 30,589 in 2019Q4, reflecting the growing volume and breadth of safety data. In parallel, the number of positive controls in the reference set expanded from 603 to 1,948. However, the proportion of positives decreased over time due to the removal of events once added to product labels.
	
	Figure \ref{fig:fig_01} shows that IC SSM consistently outperformed IC HLGT and exceeded standard IC during the early quarters of follow-up, up to 2018Q2. Although trade-offs between sensitivity and positive PPV remained, the trajectory of F1 scores (Supplementary Figure \ref{fig:fig_s01}) demonstrated that IC SSM evolved from parity with the comparators to a stable, intermediate position -- superior to IC HLGT and competitive with standard IC after 2017Q1. These results highlight the method's strength in early detection, even when aggregate performance metrics such as F1 remain modest.
	
	\subsection{Performance analyses over the entire study period}
	
\begin{figure}
		\centering
		\fbox{\includegraphics[width=0.95\textwidth]{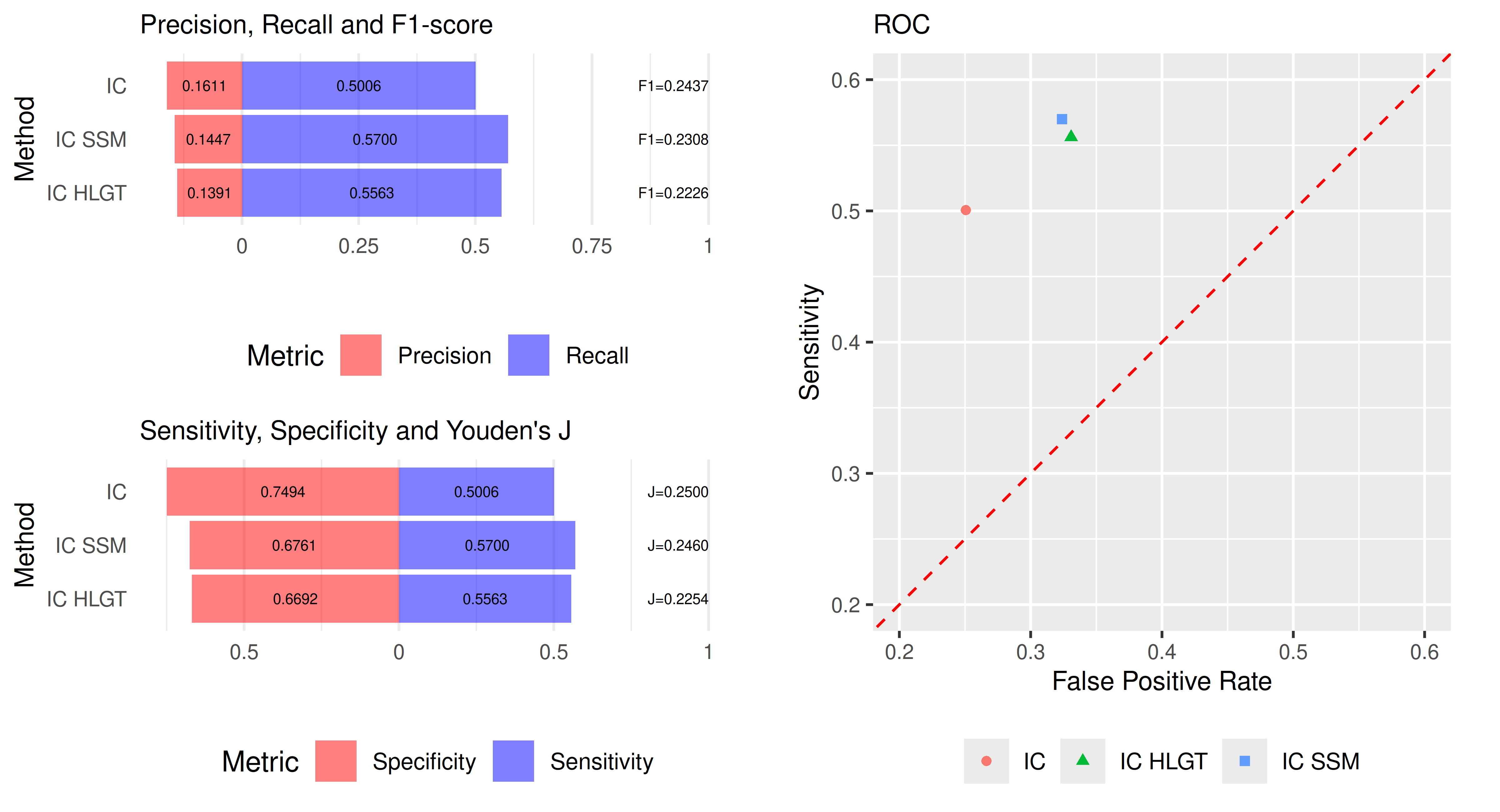}}
		\caption{Performance analyses over the entire study period}
		\label{fig:fig_02}
	\end{figure}

	Performance analyses over the entire study period are presented in figure \ref{fig:fig_02}. The results revealed important differences in various metrics when comparing IC analysis without borrowing to IC SSM. specifically, the sensitivity of IC analysis without borrowing was 50.1\%, whereas IC SSM exhibited an increased sensitivity of 57.0\%. This increase in sensitivity was accompanied by a decrease in precision, from 16.1\% to 14.5\%, resulting in a decrease in the F1 score from 0.244 to 0.231. Additionally, specificity decreased from 74.9\% with IC analysis to 67.6\% with IC SSM. The resulting Youden's index was slightly lower for IC SSM than for IC without borrowing, with 0.246 and 0.250, respectively. Comparing IC HLGT to IC SSM, the differences were less pronounced than those between IC and IC SSM. However, IC HLGT exhibited lower sensitivity, specificity, and precision, resulting in a lower F1 score (0.223) and Youden's index (0.225). This positioned IC SSM as a more effective approach than IC HLGT, and potentially also than IC, given the importance of sensitivity in quantitative signal detection for safety. When precision or specificity needed to be improved, for instance, to consider the limits of resources for the signal management process, the threshold of the lower limit of the 95\% CI could be increased. Supplementary figure \ref{fig:fig_s02} shows that with an increased threshold, the F1 score and Youden's index were higher for IC SSM compared to IC, with F1 scores of 0.239 and 0.236, and Youden's indexes of 0.191 and 0.181, respectively, when the threshold was set to 1.
	
\begin{figure}
		\centering
		\fbox{\includegraphics[width=0.95\textwidth]{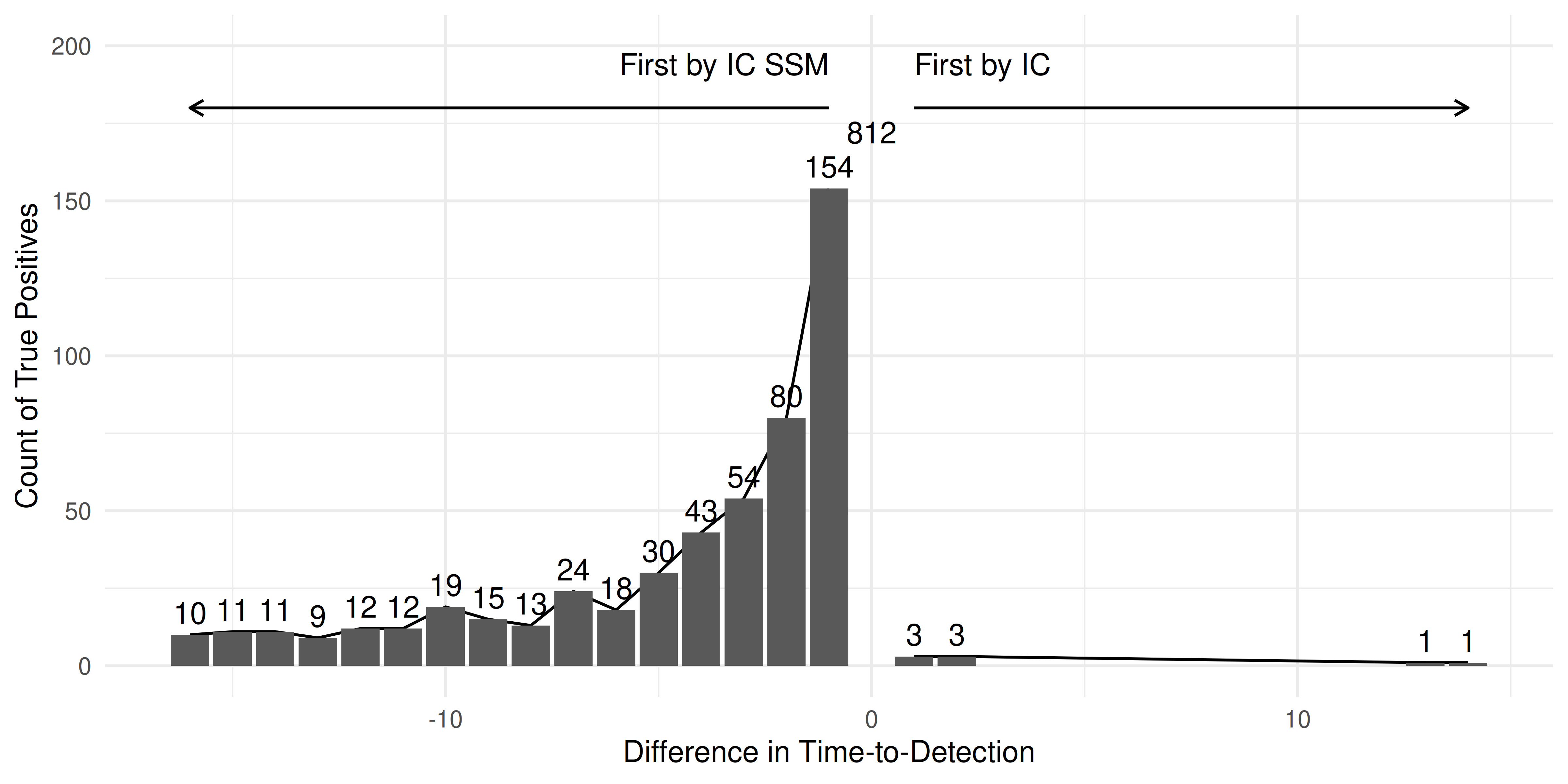}}
		\caption{Distribution of the difference in time-to-detection of true positives between IC SSM and IC}
		\label{fig:fig_03}
	\end{figure}

\begin{figure}
		\centering
		\fbox{\includegraphics[width=0.95\textwidth]{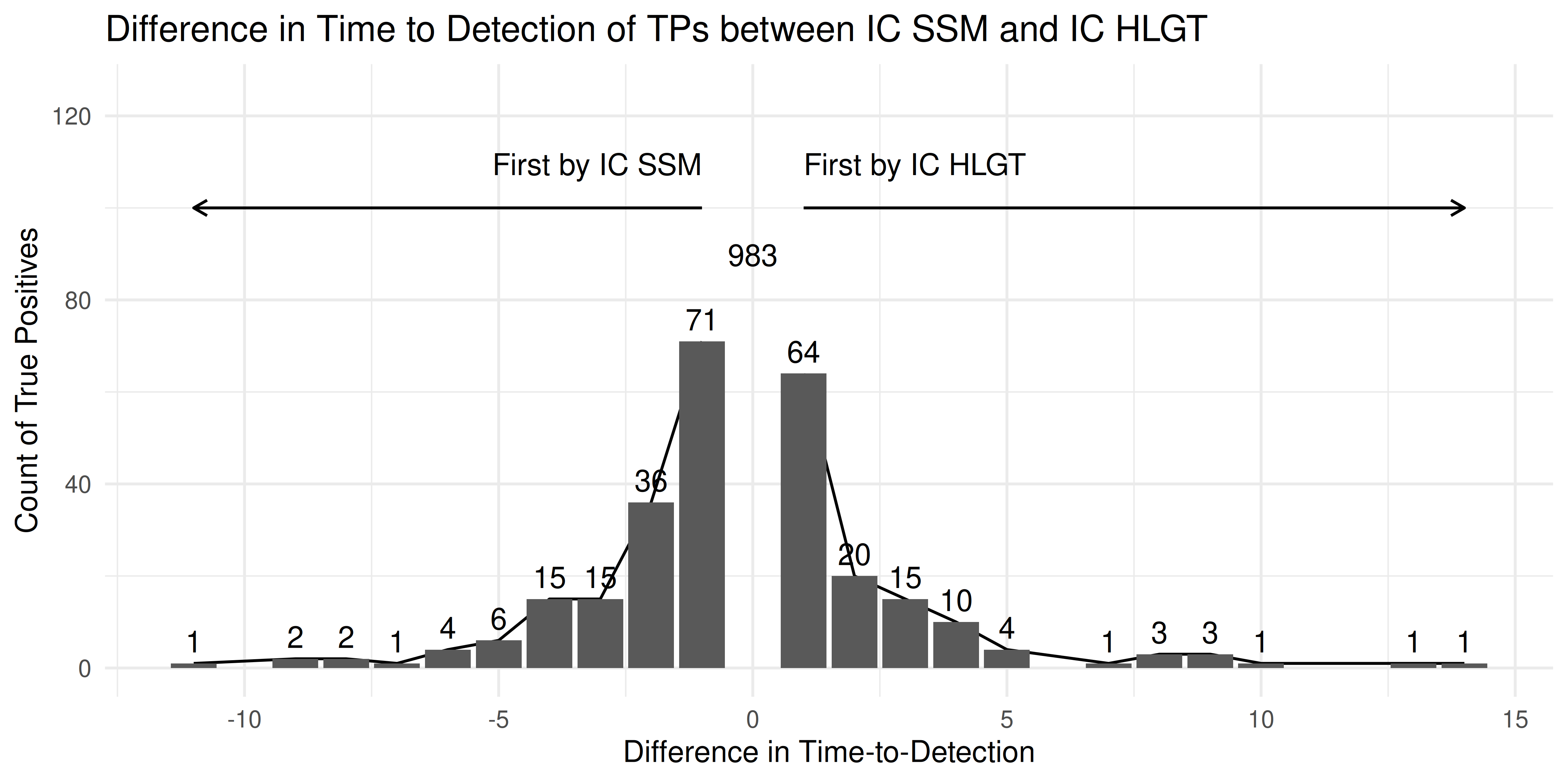}}
		\caption{Distribution of the difference in time-to-detection of true positives between IC SSM and IC HLGT}
		\label{fig:fig_04}
	\end{figure}

	Two-by-two comparisons between methods are summarized in Table \ref{tab:tbt_pos_cont} and figures \ref{fig:fig_03} and \ref{fig:fig_04}. These comparisons focus on the absolute numbers of true positives that were concordant and discordant between each method, providing a more detailed view of their relative performance. IC SSM detected 165 positive controls that IC analysis did not identify, while IC analysis detected 3 positive controls that IC SSM missed. Among the 1,167 positive controls detected by both methods, 812 were identified within the same calendar quarter. The distribution of differences in time-to-detection of positive controls is depicted in figure \ref{fig:fig_03}, illustrating that IC SSM generally detected more positive controls and earlier than IC analysis, with an average difference of 1.75 quarters. When comparing IC SSM with IC HLGT, IC SSM identified 73 positive controls that IC HLGT did not detect, while IC HLGT found 41 positive controls that IC SSM missed. Of the 1,259 positive controls detected by both methods, 983 were detected in the same calendar quarter. The distribution of differences in time-to-detection of positive controls favored IC SSM. 
	
	We assessed the robustness of IC SSM’s superior performance across variations in the reference set by employing a resampling method. New reference sets of equivalent size were generated through random sampling with replacement, and performance metrics were computed for each set. The results indicated that IC SSM demonstrated higher sensitivity compared to IC HLGT in 99.9\% of the simulations. Furthermore, IC SSM exhibited superior performance in terms of Youden's index, specificity, and F1 score in 100\% of the simulations when compared to IC HLGT. Additionally, IC SSM achieved a lower F1 score than the IC without Bayesian borrowing in 100\% of the simulations, and it demonstrated a higher Youden's index than the IC without Bayesian borrowing in 24.2\% of the simulations. 	
	
	\subsection{Parameter Sensitivity Analysis}
	
	The performance of IC SSM is influenced by three main parameters: the prior value of the weight of the MAP prior ($w$), the standard deviation of the vague prior ($\sigma$), and the lower limit of SSM (minSSM) for PTs to be included in the MAP prior according to SSM.
	
	Supplementary figure \ref{fig:fig_s03} shows the performance metrics of IC SSM for varying minSSM thresholds, with $w$ and $\sigma$ kept constant. Thresholds of 0.1 to 0.6 were considered. Sensitivity increased from 56.05\% to 57.04\% as the threshold decreased from 0.6 to 0.4, with specificity decreasing from 68.8\% to 67.9\%. The highest F1 score (0.233) and Youden’s index (0.249) were achieved with thresholds of 0.6 and 0.4, respectively. IC SSM's F1 scores and Youden’s indexes were lower than IC without borrowing but higher than IC HLGT.
	
	Supplementary figure \ref{fig:fig_s04} summarizes the impact of varying $w$ from 70\% to 90\%. Sensitivity increased while specificity and precision decreased with increasing $w$ for both IC SSM and IC HLGT. The highest Youden's index for IC SSM was 0.2394 at w=80\%. IC SSM's F1-score and Youden's index were superior to IC HLGT and inferior to IC.
	
	Supplementary figure \ref{fig:fig_s05} summarizes the impact of varying $\sigma$ from 0.1 to 10. Sensitivity increased while precision and specificity decreased with increasing $\sigma$ for both IC SSM and IC HLGT. The highest Youden’s index for IC SSM was 0.259 at $\sigma = 0.5$, followed by 0.254 at $\sigma = 1$, and the simple IC positioned itself between IC SSM with $\sigma = 1$ and $\sigma = 2$. IC SSM demonstrated higher F1 scores and Youden’s indexes than IC HLGT for the same $\sigma$ values, with differences in sensitivity increasing and differences in specificity decreasing with increasing $\sigma$.
	
	\subsection{Examples of Product-event Pairs Showing the benefits of BDB Using SSM}
	
	Figures \ref{fig:fig_05}, \ref{fig:fig_06}, and \ref{fig:fig_07} present three examples of distinct product-event pairs\footnote{In each example, the products are distinct from one another.} in which the IC SSM method outperformed both the simple IC analysis and IC HLGT in detecting AE signals.
	
	\subsubsection{Example 1: Meningitis}
	
\begin{figure}[thp]
		\centering
		\fbox{\includegraphics[width=0.95\textwidth]{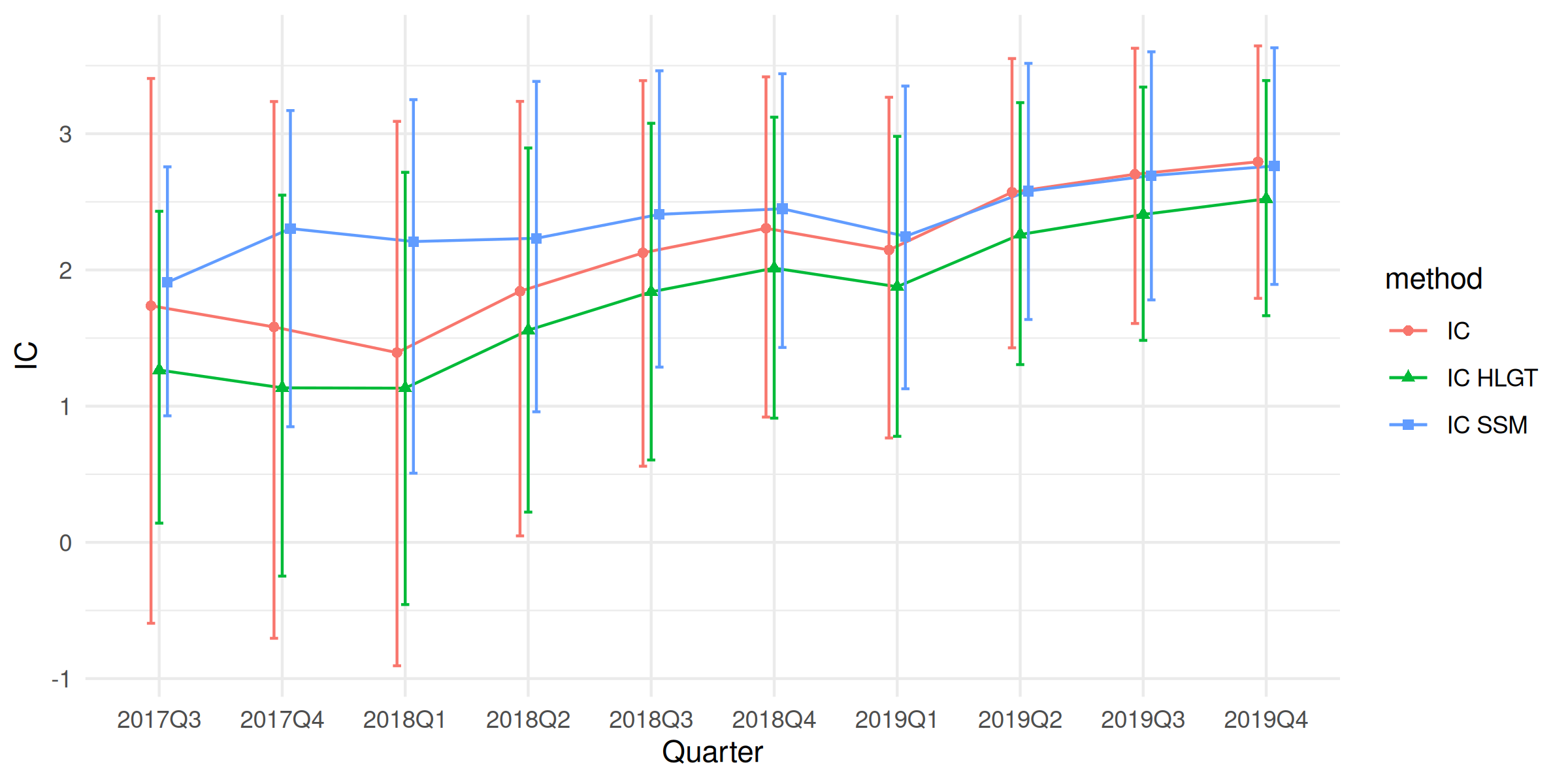}}
		\caption{Quarterly analyses example 1: Meningitis}
		\label{fig:fig_05}
	\end{figure}

	In this example (figure \ref{fig:fig_05}), the posterior medians of all three methods -- IC SSM, IC HLGT, and IC -- converged as more data accrued over time. However, IC HLGT consistently lagged behind the other two in magnitude. Notably, by 2019Q1, IC SSM provided higher and more stable point estimates compared to the other methods. Moreover, IC SSM consistently demonstrated narrower 95\% credibility intervals (CI), with lower limits that remained above zero across all quarters. This was especially prominent during the first three quarters, when the lower limits of the simple IC method fell below zero. The lower limit of IC HLGT also dropped below zero during the second and third quarters, following an initially positive value.
	
	The HLGT category for the PT code \textit{Meningitis} (Code: 10027199) -- `Infections - pathogen unspecified' (Code: 10021879) -- is broad. Examination of other PTs contributing to IC HLGT showed that several terms within the same HLGT, such as \textit{Pneumonia} (Code: 10035664), \textit{Nasopharyngitis} (Code: 10028810), and \textit{Respiratory tract infection} (Code: 10062352), introduced bias and diluted the signal. In contrast, the SSM-weighted borrowing in IC SSM prioritized more clinically relevant PTs, including \textit{Meningitis bacterial} (Code: 10027202, Sokal = 0.87) and \textit{Meningitis aseptic} (Code: 10027201, Sokal = 0.86), which contributed more accurately to the signal.
	
	\subsubsection{Example 2: Mouth Ulceration}
	
\begin{figure}[h]
		\centering
		\fbox{\includegraphics[width=0.95\textwidth]{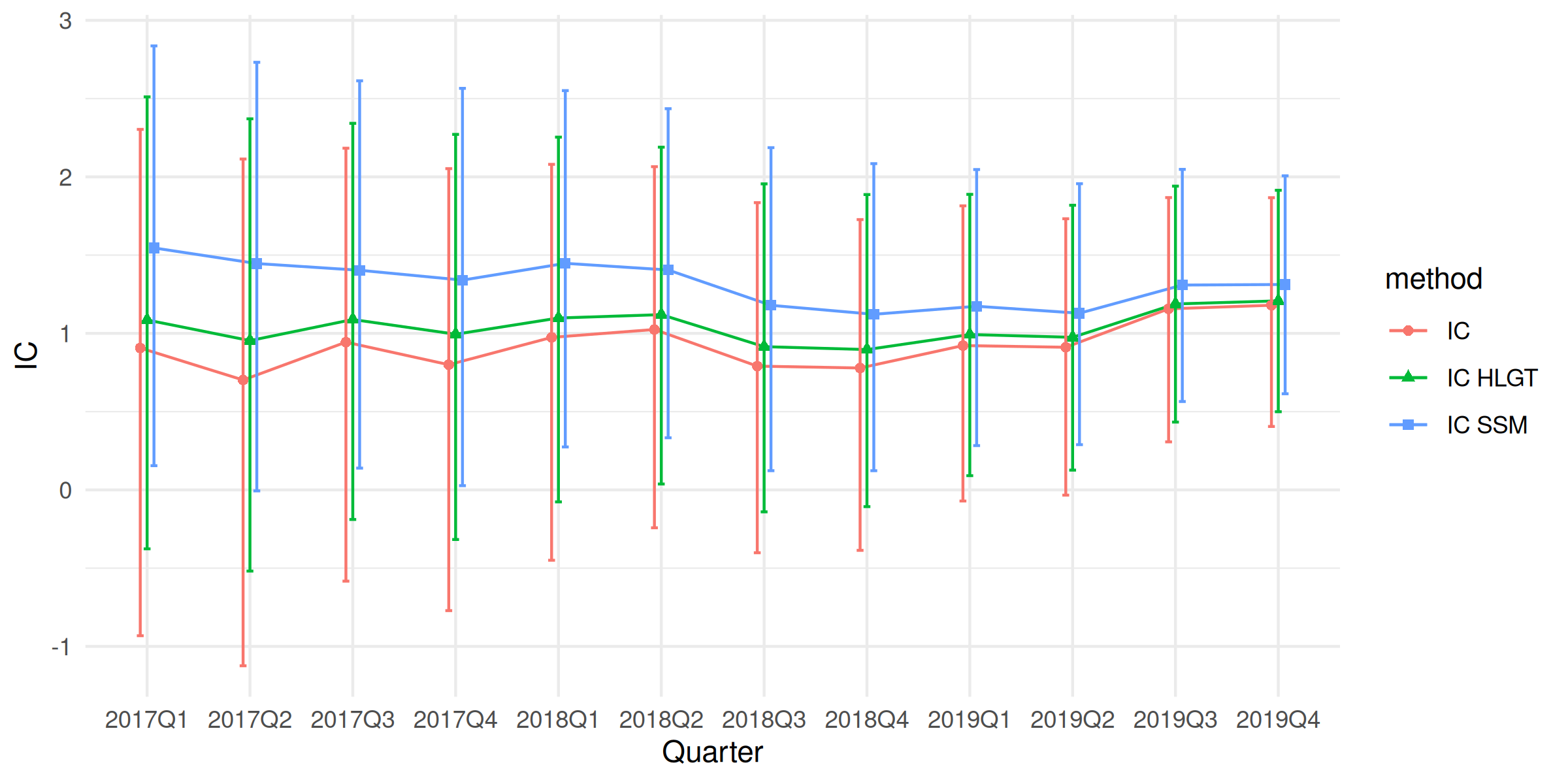}}
		\caption{Quarterly analyses example 2: Mouth Ulceration}
		\label{fig:fig_06}
	\end{figure}

	In this example (figure \ref{fig:fig_06}), the IC SSM method consistently produced higher point estimates and tighter lower credibility bounds compared to both the standard IC and IC HLGT approaches. IC SSM identified the AE five quarters earlier than IC HLGT and approximately 2.5 years sooner than the simple IC method. 
	
	Within the HLGT \textit{Oral soft tissue conditions} (Code: 10031013), the most influential MedDRA PT in the MAP prior under IC HLGT was \textit{Stomatitis} (Code: 10042128), which contributed 61 cases by 2017Q1 and had a 95\% credibility interval (CI) of $[3.44,4.15]$. This was followed by \textit{Oral pain} (Code: 10031009), with 38 cases and a CI of $[3.64,4.55]$. While both terms are clinically relevant to \textit{Mouth ulceration} (Code: 10028034), they are broader and less specific.
	
	In the IC SSM model, \textit{Stomatitis} also carried the greatest weight, but its influence was moderated due to a Sokal semantic similarity score of $0.63$ -- only moderately close to \textit{Mouth ulceration}. The next most influential PT in the IC SSM MAP prior was \textit{Mucosal inflammation} (Code: 10028116), from a different HLGT -- \textit{General system disorders NEC} (Code: 10018073) -- with 8 cases and a 95\% CI of $[1.08,3.11]$ and a Sokal similarity of 0.51. In total, three other PTs with Sokal scores above $0.3$ exhibited IC values more consistent with \textit{Mouth ulceration}, collectively contributing to a more targeted and effective borrowing structure. The PT with the highest similarity was \textit{Aphthous ulcer} (Code: 10002959), with a Sokal similarity of $0.96$. 
	
	These results highlight how the IC SSM method selectively draws information from clinically similar terms across HLGT boundaries. In addition, the PTs included in the MAP prior of the IC SSM had 95\% CI of IC much more consistent with \textit{Mouth ulceration}, leading to a higher weight of the MAP prior in the results thanks to the dynamic borrowing, potentially explaining the more significant shrinkage, leading to improved early detection and more stable estimates. 			
	
	\subsubsection{Example 3: Oedema Peripheral}
	
\begin{figure}[h]
		\centering
		\fbox{\includegraphics[width=0.95\textwidth]{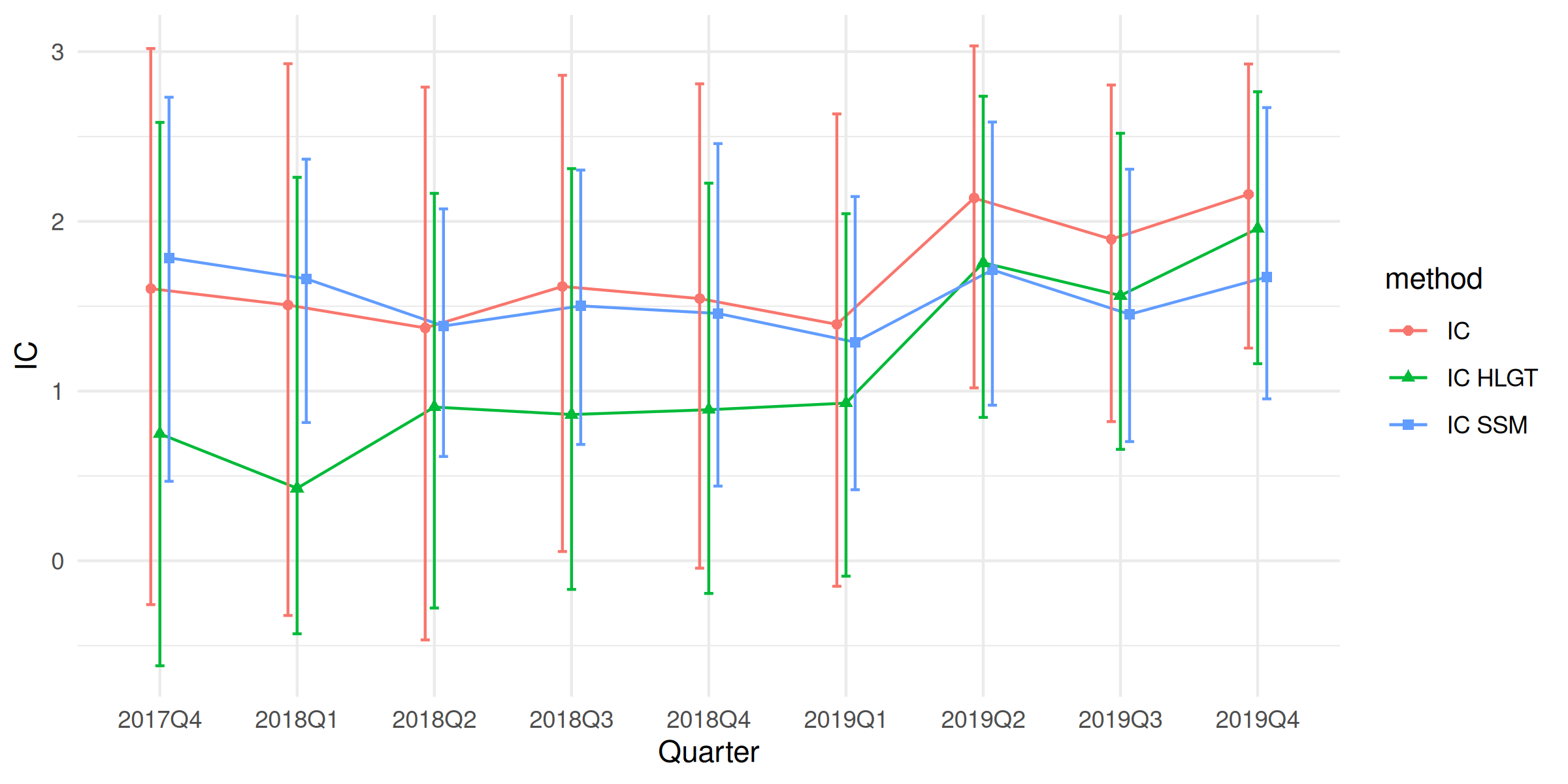}}
		\caption{Quarterly analyses example 3: Oedema Peripheral}
		\label{fig:fig_07}
	\end{figure}

	This example (figure \ref{fig:fig_07}) demonstrates a case where borrowing from the HLGT level was detrimental to performance. During the first six quarters, the point estimates for both IC and IC SSM were approximately 1.5, while IC HLGT remained below 1. Notably, the lower bound of the IC SSM 95\% credibility interval remained consistently above zero, supporting its signal stability. In contrast, the simple IC method produced a weak and transient signal in the fourth quarter that did not persist.
	
	The most pronounced divergence between IC SSM and IC HLGT occurred in the second quarter. At that time, only three PTs from the HLGT \textit{General system disorders NEC} (Code: 10018073) contributed data. All exhibited lower IC estimates and small case counts. The most influential of these was \textit{Fatigue} (Code: 10016256), with only 6 cases and a wide CI of $[-1.20,\ 1.04]$.
	
	In contrast, IC SSM incorporated borrowing from two PTs with higher semantic similarity and more consistent IC values: \textit{Pulmonary oedema} (Code: 10037423, Sokal = 0.60, CI = $[0.17,\ 3.40]$) and \textit{Cardiac failure} (Code: 10007554, Sokal = 0.81, CI = $[-1.21,\ 2.74]$). These terms contributed meaningfully to the MAP prior in the IC SSM model, helping to sustain a stronger and earlier signal. This example reinforces the value of selective borrowing guided by semantic similarity and highlights the limitations of HLGT-based aggregation when component PTs are weakly related to the target.

	\section{Discussion}
	
	This study introduced and evaluated a BDB method that integrates SSMs into DPA for SRSs of AEs. Using a robust meta-analytic predictive (MAP) prior within a Bayesian hierarchical model, the proposed approach enables borrowing from MedDRA PTs based on their semantic similarity to the PT of interest. PTs with SSM values greater than 0.3 were included in the MAP prior, weighted by their SSM value divided by the estimated IC variance. A robustifying prior (Normal with mean = 0, SD = 2) was applied, and the maximum SSM was used to define the weight assigned to the MAP prior. Performance was compared to traditional IC analysis and IC analysis with BDB at the MedDRA HLGT level, using FAERS data. Evaluation relied on an automated, time-stamped reference set derived from FDA safety labeling updates.
	
	The IC SSM approach demonstrated improved sensitivity compared to both traditional IC and HLGT-based borrowing, with minor trade-offs in F1 score and Youden's index. While IC SSM outperformed HLGT-based borrowing in all performance analyses, the differences were incremental compared to those between IC SSM and traditional IC. IC SSM consistently identified more true positives and detected signals earlier than other methods, averaging over 5 months sooner than traditional IC. Although its aggregate Youden index was marginally lower than that of traditional IC, quarterly analyses showed that IC SSM had higher performance in the early post-marketing period -- a critical window for identifying safety signals before widespread exposure. These observations were reinforced by detailed case examples, where SSM-based borrowing provided more stable and contextually relevant estimates than HLGT-based borrowing. 
	
	Where greater specificity or precision is desired—such as under constrained signal management resources—raising the threshold on the lower bound of the 95\% credibility interval offers a practical adjustment. While sensitivity declines with this change, IC SSM maintained higher F1 scores and Youden's indices compared to IC across several thresholds. Notably, IC SSM surpassed IC in Youden's index when the signal threshold was increased to 0.2, suggesting it offers a more favorable balance between sensitivity and specificity.
	
	The benefits of IC SSM were clearest in cases where the PT of interest belonged to broad or clinically diffuse HLGTs such as \textit{Infections - pathogen unspecified} (Code: 10021879) or \textit{General system disorders NEC} (Code: 10018073). In these scenarios, SSMs enabled selective borrowing from PTs outside the HLGT that were clinically relevant and statistically informative. For instance, in the analysis of \textit{Mouth ulceration}, SSM-based borrowing included PTs with IC values more aligned with the target PT than those included under HLGT grouping. These findings underscore how IC SSM adapts borrowing based on semantic proximity, supporting more accurate and stable estimation.

	Sensitivity analyses further clarified the effects of tuning key parameters: the prior weight $w$ and the standard deviation of the robustifying prior $\sigma$. Increasing $w$ boosted sensitivity but decreased specificity and precision for both IC SSM and IC HLGT. While F1 scores declined at higher weights, the configuration using $\max(\mathrm{SSM})$ -- our reference setting -- produced the highest F1 score overall. Across all $w$ values, IC SSM outperformed IC HLGT, though traditional IC retained the highest F1. Similarly, decreasing $\sigma$ increased sensitivity but reduced specificity. IC SSM consistently outperformed IC HLGT in both F1 score and Youden's index across all $\sigma$ values, and achieved higher Youden's index than traditional IC at $\sigma < 2$, where the influence of similar PTs is maximized and posterior uncertainty is minimized. While traditional IC still achieved the highest F1 score, it did not outperform IC SSM in early detection or in the sensitivity-specificity trade-off, suggesting that performance differences persist even when tuning parameters to optimize equivalence. 		
	
	Across all analyses, variation in sensitivity had a larger impact on overall performance than the relatively narrow range of PPVs. Given the high stakes of missing true safety signals in PV, sensitivity is often prioritized over PPV. While we did not calculate F-beta scores, which assign greater weight to sensitivity, IC SSM would likely show even stronger relative performance under such metrics.
	
	These results are broadly consistent with previous findings. The IMI-PROTECT study \cite{RN48} reported a mean sensitivity of 0.30 for IC—substantially lower than our results for both IC (0.50) and IC SSM (0.57). Their slightly higher precision (0.18, versus 0.14 for IC SSM) resulted in a marginally lower F1 score (0.225 vs. 0.244 and 0.231 for our IC and IC SSM, respectively). These differences likely reflect the use of older data, other SRS sources, and a smaller, non-time-stamped reference set. Because their study did not evaluate the detection of unknown AEs, comparability is limited. Evaluation of vigiRank \cite{RN49}, which combines DPA with AE case-level features, reported even higher performance (sensitivity $>$ 60\%, specificity $>$ 70\%, AUC $>$ 65\%). However, these results are not directly comparable due to differences in methodology, reference sets, and data sources, underscoring the value of evaluating methods within a single consistent setting.
	
	There are important limitations to our work. While our model draws inspiration from meta-analytic re-weighting strategies based on quality scores - some of which have been criticized for producing biased effect estimates due to the non-independence of quality scores from study features \cite{RN50} -- our use of ontology-based SSMs ensures that MedDRA PTs are weighted independently of their potential reporting biases. Model parameters were selected empirically, and potential confounders such as age, sex, and report year were not adjusted due to inconsistent reporting and to the short period of analysis. Semantic similarity captures clinical relatedness but may not fully reflect spontaneous reporting patterns, which are influenced by patient behavior, media coverage, and reporting incentives. Finally, our analyses used all available data for parameter selection and model tuning. Although no external validation set was used, we performed sensitivity analyses using alternate reference sets to evaluate robustness.
	
	Despite these limitations, this study has several strengths. We employed a large, time-stamped reference set derived from regulatory labeling changes \cite{RN51, RN46}, enabling prospective evaluation of detection performance. The IC SSM framework builds on an established Bayesian method that is computationally efficient and scalable to large datasets. It also addresses a fundamental challenge in PV: defining meaningful groupings of medical concepts. Unlike rule-based groupings in MedDRA (e.g., SMQs or HLGTs), IC SSM uses continuous, data-driven similarity weights that better reflect the nuanced relationships between clinical events.
	
	To our knowledge, this is the first implementation of SSM-informed borrowing in DPA for AE signal detection in spontaneous reporting. As a hypothesis generation tool, it offers practical benefits for early detection of safety concerns. Future work could expand on this foundation by exploring alternative borrowing models, incorporating additional or transformed similarity metrics, or applying Bayesian inference directly to case-level data. Novel SSMs could be developed to better reflect spontaneous reporting dynamics—for instance, by integrating co-reporting patterns or biological pathways. Testing IC SSM with vigiVec-derived similarities \cite{RN24} or comparing it directly to vigiRank \cite{RN49} may yield further insight. Moreover, combining exposure-based and outcome-based BDB methods may enhance signal detection. Broader application to other real-world data sources—such as claims or EHR data—will help assess generalizability across diverse PV settings.

	\section{Conclusion}
	
	This study shows that semantic similarity-informed borrowing can improve post-marketing safety surveillance by enabling more nuanced disproportionality analysis (DPA).  Compared to the standard IC approach -- routinely used in PV -- and borrowing at the MedDRA HLGT level, the IC SSM method has the ability to detect true positive signals earlier, with improved sensitivity and favorable trade-offs in F1 score and Youden's index.
	
	While spontaneous reporting systems such as FAERS are known to contain incomplete or inconsistent data, particularly in how events are coded and reported, our results suggest that leveraging semantic similarity between MedDRA terms offers a more robust and clinically meaningful basis for information sharing than rigid HLGT-based inclusion criteria. Rather than relying on binary membership in predefined groupings, SSM allows for continuous, graded borrowing that reflects the spectrum of clinical similarity between events.
	
	To our knowledge, this is the first implementation of SSM-informed borrowing in DPA for AE signal detection in spontaneous reporting databases. Future research should aim to replicate these results in other datasets and regulatory contexts, and explore alternative Bayesian borrowing strategies, novel similarity measures, and integration with case-level or real-world data sources.
	
	\section*{Declarations}
	\textbf{Author contributions} \\
	
	François Haguinet and Jeffery Painter contributed to the conceptual design of the study, authored the manuscript, and participated in its critical revision. Gregory Powell provided clinical expertise and reviewed and edited the manuscript. Andrea Callegaro contributed to the statistical methodology and reviewed and edited the manuscript. Andrew Bate brought forward the idea of the new approach, contributed to the conceptual design, and reviewed and edited the manuscript.\\
	
	\textbf{Financial disclosure} \\
	
	GSK covered all costs associated with the conduct of the study and the development of the manuscript and the decision to publish the manuscript. J.P., F.H., G.P, A.C., and A.B. are employed by GSK and hold financial equities. 
	
	\bibliography{bbr}	
	
	\clearpage
	
	\begin{table}[h]
		\centering
		\begin{tabular}{lll}
			\toprule
			& $y$     & not $y$ \\
			\midrule
			$x$     & $a$     & $b$ \\
			not $x$ & $c$     & $d$ \\
			\bottomrule
		\end{tabular}
		\caption{Contingency table for product–event pairs}
		\label{tab:cont}
	\end{table}
	
	\clearpage
	\begin{table}[h]
		\centering
		\begin{tabular}{c ccc ccc}
			\toprule
			& \multicolumn{2}{c}{\textbf{IC}} & & \multicolumn{2}{c}{\textbf{IC HLGT}} \\
			\cmidrule(lr){2-3} \cmidrule(lr){5-6}
			\textbf{IC SSM} & $+$ & $-$ & & $+$ & $-$ \\
			\midrule
			$+$ & 1,167 & 165 & & 1,259 & 73 \\
			$-$ & 3 & 1,002 & & 41 & 964 \\
			\bottomrule
		\end{tabular}
		\caption{Two-by-two contingency tables comparing IC SSM with IC and IC HLGT for detecting positive controls}
		\label{tab:tbt_pos_cont}
	\end{table}
	
	\clearpage
	\section*{Appendix}
	
	\setcounter{figure}{0}
	\renewcommand{\thefigure}{S\arabic{figure}}
	
	\begin{figure}[h]
		\centering
		\fbox{\includegraphics[width=0.95\textwidth]{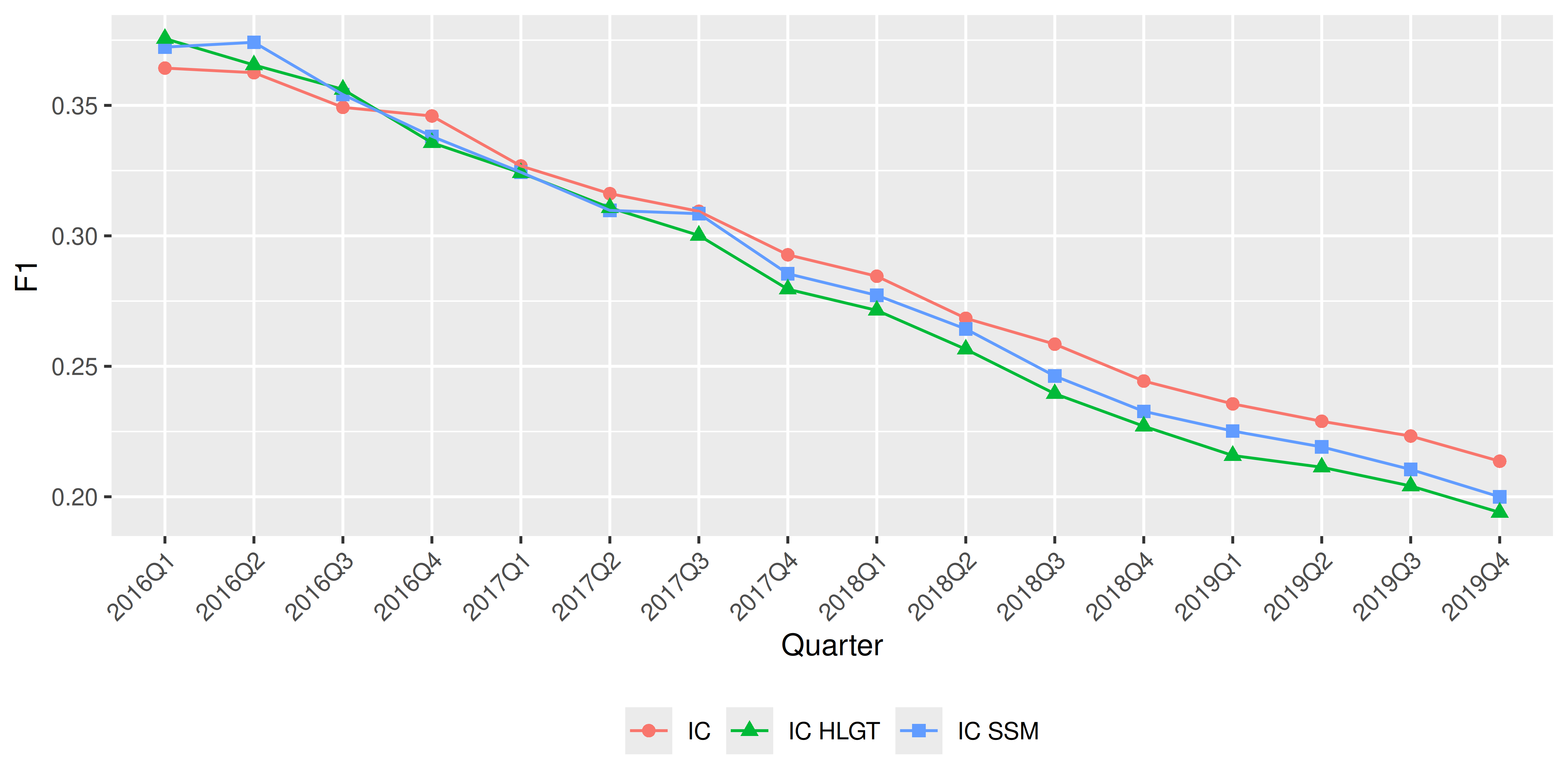}}
		\caption{Quarterly evolution of the F1 score for the three methods. IC SSM generally tracked between IC and IC HLGT throughout the period, demonstrating modest but consistent improvements over HLGT-based borrowing, particularly after 2017Q1. All methods show a declining trend over time as the reference set grows and the proportion of new positive controls decreases.}
		\label{fig:fig_s01}
	\end{figure}
	
	\begin{figure}
		\centering
		\fbox{\includegraphics[width=0.95\textwidth]{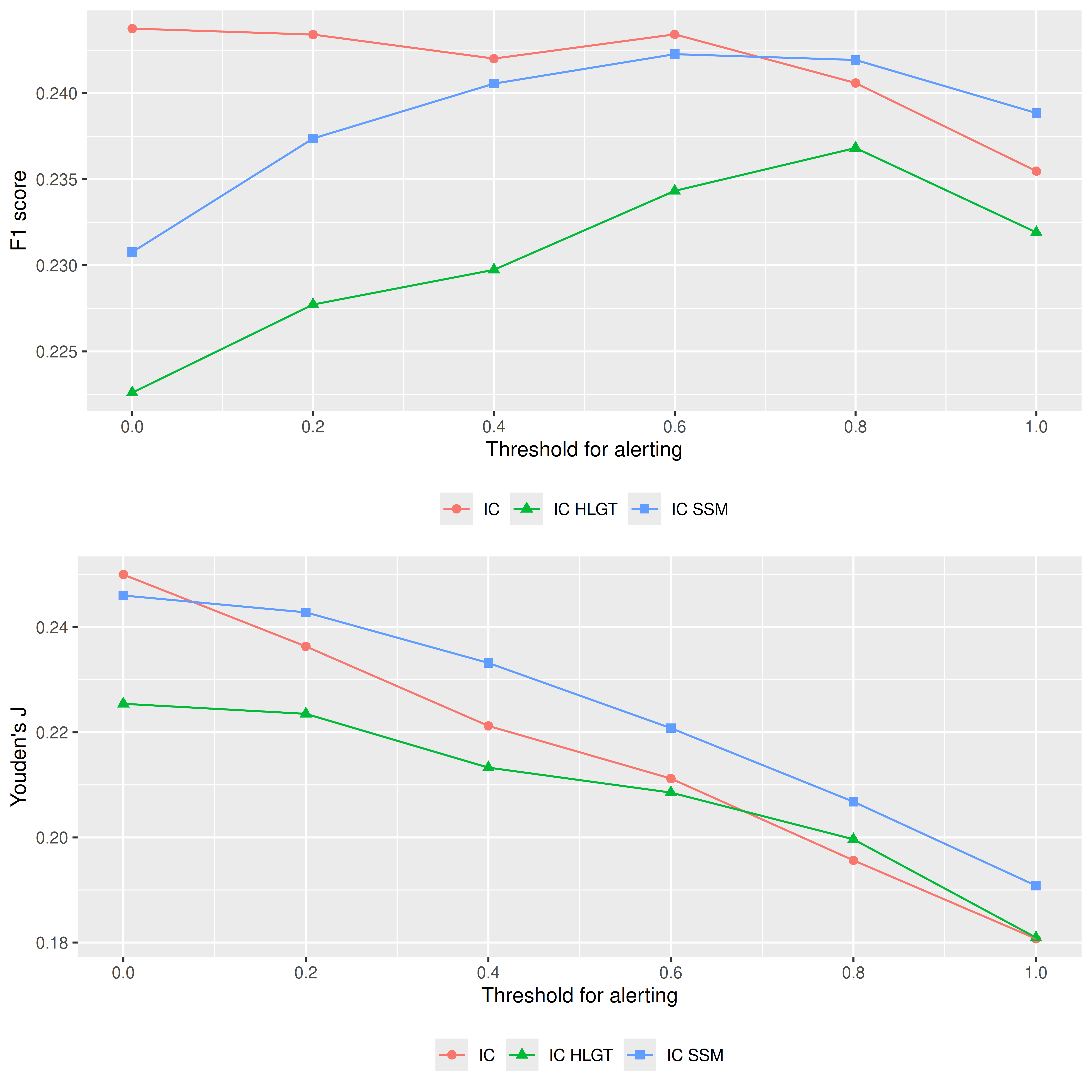}}
		\caption{Performances for varying thresholds of the lower limit of the 95\% CI of IC}
		\label{fig:fig_s02}
	\end{figure}

	\begin{figure}
		\centering
		\fbox{\includegraphics[width=0.95\textwidth]{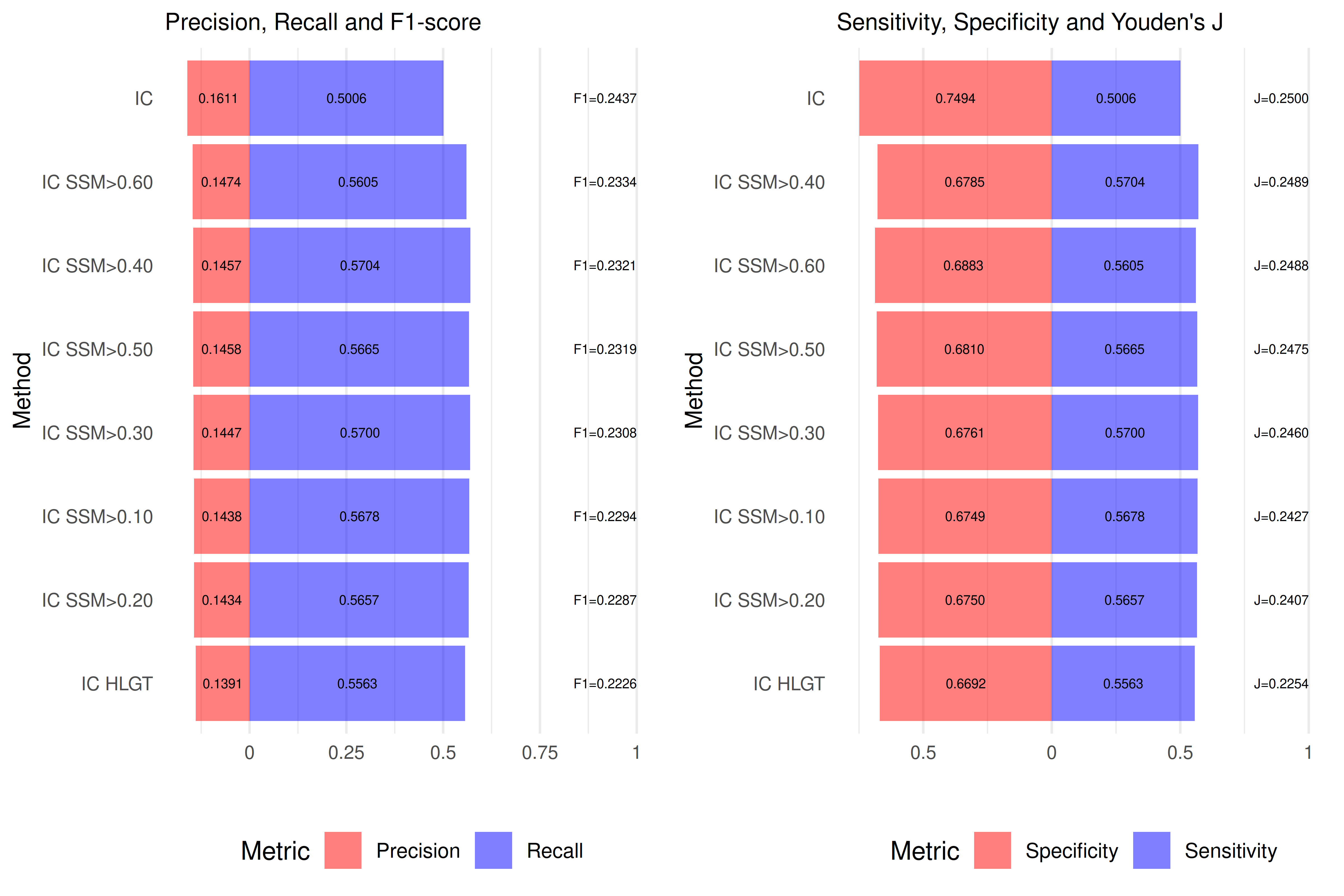}}
		\caption{IC SSM performances for varying thresholds of minimum SSM for inclusion in MAP prior}
		\label{fig:fig_s03}
	\end{figure}

	\begin{figure}
		\centering
		\fbox{\includegraphics[width=0.95\textwidth]{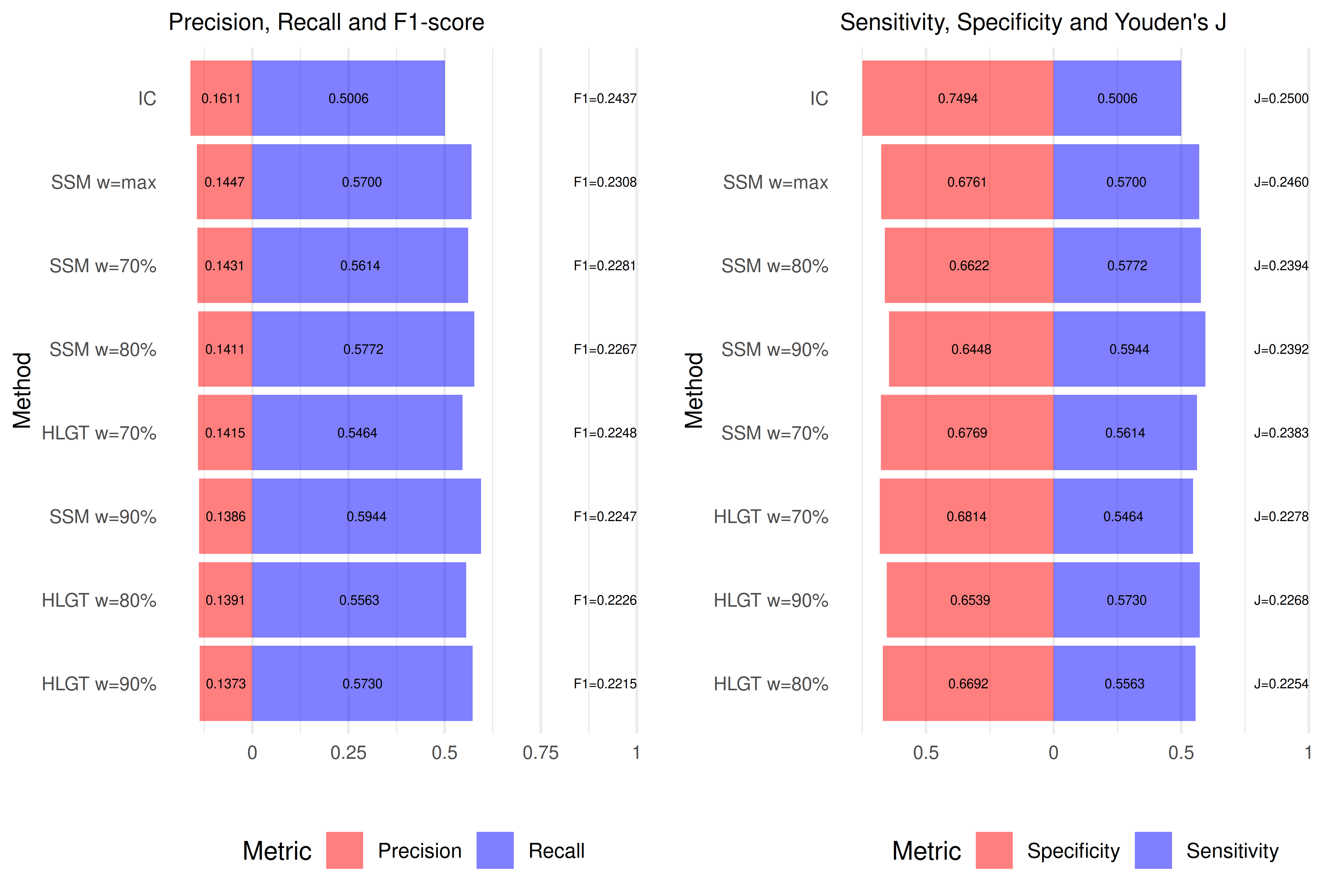}}
		\caption{IC SSM performances for varying weight of the MAP prior in the robust MAP prior }
		\label{fig:fig_s04}
	\end{figure}

	\begin{figure}
		\centering
		\fbox{\includegraphics[width=0.95\textwidth]{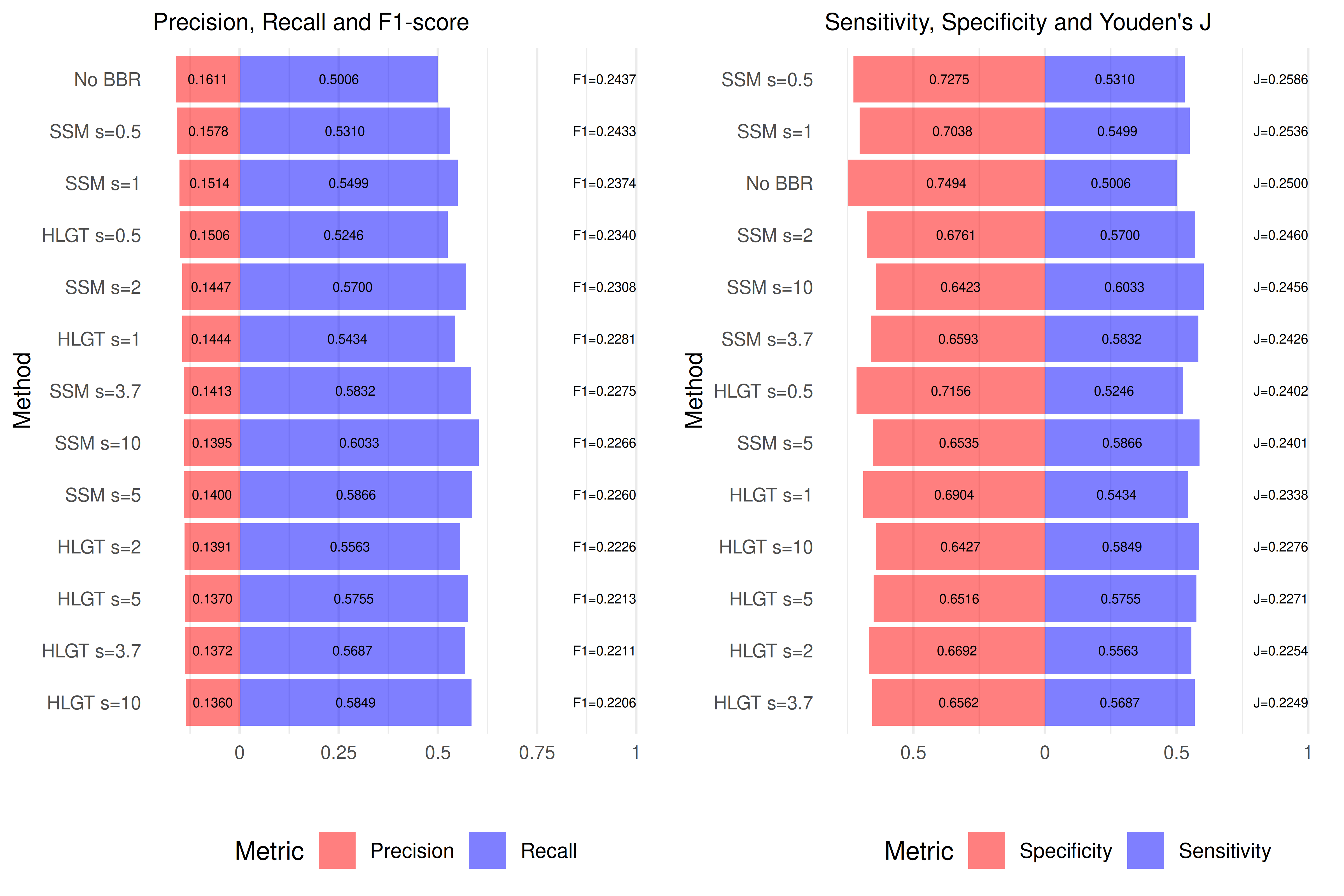}}
		\caption{Performances for varying $\sigma$ in the vague prior ($s = \sigma$). No BBR is the traditional IC.}
		\label{fig:fig_s05}
	\end{figure}

\end{document}